\documentclass[letterpaper, 10pt, conference]{ieeeconf}

\IEEEoverridecommandlockouts
\usepackage[dvipsnames]{xcolor}
\usepackage{amsmath}
\usepackage{amssymb}
\usepackage{bm}
\usepackage{booktabs}
\usepackage{caption}
\usepackage{csquotes}
\usepackage{epsfig}
\usepackage{graphicx}
\usepackage{multirow}
\usepackage{tabularx}
\usepackage{times}
\usepackage{xspace}
\usepackage{float}

\usepackage[pagebackref=true,breaklinks=true,letterpaper=true,colorlinks,bookmarks=false]{hyperref}
\usepackage[capitalize]{cleveref}

\newtoggle{nographics}
\newtoggle{lrgraphics}
\newtoggle{nocomments}
\newtoggle{supplementary}

\togglefalse{nographics}
\togglefalse{lrgraphics}
\togglefalse{nocomments}
\togglefalse{supplementary}

\crefname{section}{Section}{Sections}
\crefname{table}{Table}{Tables}
\crefname{figure}{Figure}{Figures}

\newcommand{\Kill}[1]{}
\iftoggle{nocomments}{
    \newcommand{\LA}[1]{\ignorespaces}
    \newcommand{\AK}[1]{\ignorespaces}
    \newcommand{\KY}[1]{\ignorespaces}
    \newcommand{\todo}[1]{\ignorespaces}
    \newcommand{\plan}[1]{\ignorespaces}
}{
    \newcommand{\LA}[1]{\textbf{\textcolor{violet}{AA: #1}}}
    \newcommand{\AK}[1]{\textbf{\textcolor{orange}{AK: #1}}}
    \newcommand{\KY}[1]{\textbf{\textcolor{blue}{KY: #1}}}
    \newcommand{\todo}[1]{\textbf{\textcolor{red}{#1}}}
    \newcommand{\plan}[1]{\textsl{\textcolor{gray}{#1}}}
}

\DeclareGraphicsExtensions{.eps,.pdf,.jpg,.png}
\iftoggle{lrgraphics}{
    \graphicspath{{src/media/lr/}{src/media/vector/}{src/media/tables/}}
}{
    \graphicspath{{src/media/hr/}{src/media/vector/}{src/media/tables/}}
}
\makeatletter
\iftoggle{nographics}{
    \LetLtxMacro{\includegraphics@orig}{\includegraphics}
    \RenewDocumentCommand{\includegraphics}{ s O{} m }{%
        {\setlength{\fboxsep}{0pt}%
         \colorbox{lightgray}{\phantom{\IfBooleanTF{#1}{\includegraphics@orig*}{\includegraphics@orig}[#2]{#3}}}%
        }%
    }
}{}
\makeatother

\crefname{figure}{Fig.}{Figs.}
\Crefname{figure}{Figure}{Figures}
\crefname{section}{Sec.}{Secs.}
\Crefname{section}{Section}{Sections}
\Crefname{table}{Table}{Tables}
\crefname{table}{Tab.}{Tabs.}

\makeatletter
\DeclareRobustCommand\onedot{\futurelet\@let@token\@onedot}
\def\@onedot{\ifx\@let@token.\else.\null\fi\xspace}

\def\eg{\emph{e.g}\onedot} 
\def\ie{\emph{i.e}\onedot} 
\def\cf{\emph{c.f}\onedot} 
 
\def\wrt{w.r.t\onedot} 

\makeatother

\begin{document}

\title{\LARGE \bf
DeepMIF: Deep Monotonic Implicit Fields\\ for Large-Scale LiDAR 3D Mapping}

\author{Kutay Y{\i}lmaz$^{1}$, Matthias Nie{\ss}ner$^{1}$, Anastasiia Kornilova$^{2\text{*}}$, and Alexey Artemov$^{1\text{*}}$
\thanks{$^{1}$Kutay Y{\i}lmaz, Matthias Nie{\ss}ner, and Alexey Artemov are with the Technical University of Munich, Garching, Germany
    {E-mail: alexey.artemov@tum.de}}%
\thanks{$^{2}$Anastasiia Kornilova is with the Skolkovo Institute of Science and Technology, Moscow, Russia.}%
\thanks{$^{\text{*}}$Equal senior author contribution.}%
\thanks{Anastasiia Kornilova was supported by the Analytical center under the RF Government (subsidy agreement 000000D730321P5Q0002, Grant No. 70-2021-00145 02.11.2021).}
}

\maketitle

\begin{abstract}
Recently, significant progress has been achieved in sensing real large-scale outdoor 3D environments, particularly by using modern acquisition equipment such as LiDAR sensors.
Unfortunately, they are fundamentally limited in their ability to produce dense, complete 3D scenes.
To address this issue, recent learning-based methods integrate neural implicit representations and optimizable feature grids to approximate surfaces of 3D scenes.
However, naively fitting samples along raw LiDAR rays leads to noisy 3D mapping results due to the nature of sparse, conflicting LiDAR measurements. 
Instead, in this work we depart from fitting LiDAR data exactly, instead letting the network optimize a non-metric monotonic implicit field defined in 3D space. 
To fit our field, we design a learning system integrating a monotonicity loss that enables optimizing neural monotonic fields and leverages recent progress in large-scale 3D mapping. 
Our algorithm achieves high-quality dense 3D mapping performance as captured by multiple quantitative and perceptual measures and visual results obtained for Mai City, Newer College, and KITTI benchmarks. The code of our approach is publicly available at \href{https://github.com/artonson/deepmif}{https://github.com/artonson/deepmif}.
\end{abstract}


\section{Introduction}
\label{sec:intro}

Implicit 3D representations, \ie algorithms that represent shapes and scenes via level-sets of functions (fields) obtained by approximating sensor 3D measurements, enjoy well-deserved popularity for scene modeling. 
Their core advantages compared to other types of 3D representations (\eg, point sets or volumetric grid) consist in their ability to accurately model shapes and scenes of arbitrary topology and resolution at moderate computational cost. 
Supported by the progress in deep neural networks that can easily fuse multi-modal data, implicit 3D representations can be inferred from various types of acquisitions such as 3D samples~\cite{park2019deepsdf,mescheder2019occupancy}, RGB images~\cite{wang2021neus}, or RGB-D sequences~\cite{wang2022go,azinovic2022neural}.
As a result, neural implicit fields are starting to see interest in the robotics domain where they are explored for large-scale 3D mapping~\cite{zhong2023shine,shi2023accurate,song2024n,isaacson2023loner}, odometry estimation~\cite{deng2023nerf}, and localization~\cite{wiesmann2023locndf} applications.

\begin{figure}[t]
\centerline{
\includegraphics[
width=1\columnwidth,
trim=0 0 0 0, 
clip=True]{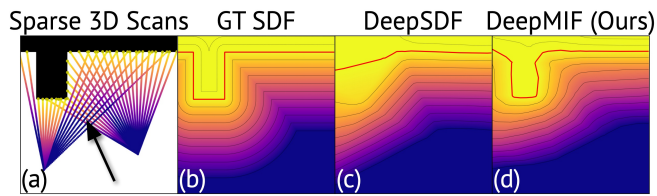}
}
\caption{LiDAR 3D scans~(a) generate view-inconsistent range data (pointed to by arrow)
rather than projective SDF~(b). 
Direct optimization supervised by oblique rather than projective distances~\cite{park2019deepsdf} does not account for this effect, resulting in loss of surface features~(c); 
in contrast, learning our implicit function~(d) is able to preserve higher detail.
Red line corresponds to zero level set.
\vspace{-1em}}
\label{fig:method-mif-vs-others}
\end{figure}

Directly extending these methods to large outdoor 3D environments typical for mobile robotics (\eg, autonomous driving) which is the focus of this work, however, is challenging.
Commonly, neural implicit models are optimized for exhaustively sampled 3D scenes and rely on accurate, consistent distance-to-surface measurements; depth cameras used in many indoor scenarios to some degree satisfy these assumptions~\cite{azinovic2022neural}.
However, sensing setups used in outdoor robotics commonly include one or many rotating LiDAR scanners; along with generating a sparse and noisy supervision signal, their acquisitions yield conflicting distance-to-surface values in arbitrarily close 3D points, due to differences in the incident angles of rays emitted from two scanning locations~\cite{deng2023nerf} (see \cref{fig:method-mif-vs-others}).
Recognizing these limitations, recent methods propose to correct distance measurements by estimating and using normals~\cite{deng2023nerf,song2024n}; however, normals estimation on noisy 3D scans can be unstable.

Motivated by these challenges, in this paper we address the case where a set of noisy, view-inconsistent LiDAR range scans is used for neural implicit surface fitting.
To this end, we propose monotonic implicit functions (MIF), scene representations suitable for addressing the challenges arising due to the view-dependent nature of LiDAR acquisitions.
Instead of requiring that accurate ground-truth measurements such as signed distance field (SDF) are produced by the LiDAR sensor, we optimize for a non-increasing (along each emitted ray) field whose zero level-set coincides with that of the true SDF, thus bypassing the need for accurate distance-to-surface values during training, but still allowing accurate surface extraction. 
To fit our field given the sensor data, we design a learning system for optimizing monotonic functions (using monotonicity loss) within a framework for fitting neural implicits integrating a hierarchical latent feature grid and adaptive point sampling.
As a result, our approach is capable of performing reconstruction of large-scale 3D LiDAR acquisitions using optimization, without any ground-truth data other than the 3D scan itself.

We evaluate our approach against five diverse methods based on depth fusion~\cite{vizzo2022vdbfusion}, interpolation~\cite{vizzo2021poisson}, completion~\cite{vizzo2022make}, and learning-based implicit reconstruction~\cite{zhong2023shine,deng2023nerf}.
We summarize our contributions as follows:
\begin{itemize}
\item We propose an alternative implicit surface representation, \textit{monotonic implicit field,} for large-scale 3D mapping with LiDAR point clouds. 
\item We demonstrate an implementation of our implicit field within a large-scale 3D mapping system that achieves improved surface reconstruction performance on multiple challenging benchmarks.
\end{itemize}

\section{Related Work}
\label{sec:related}

\begin{figure}[t]
\centerline{
\includegraphics[
width=.75\columnwidth,
trim=0 0 0 0, 
clip=True]{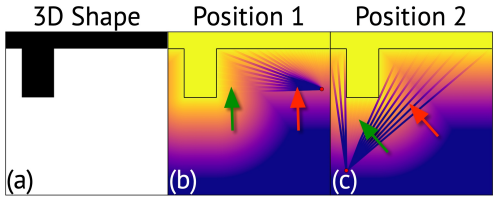}
}
\caption{LiDAR scanners generate oblique distances (distance along the ray, color of foreground lines) deviating from projective distances ((b)--(c), background color).
Depending on scanner position (red dots) and scanning angles, these quantities can be either slightly (areas pointed by \textcolor{ForestGreen}{green arrows}) or significantly (areas pointed by \textcolor{red}{red arrows}) different.
\vspace{-1em}}
\label{fig:method-oblique-sdf}
\end{figure}

3D reconstruction from scanned data has been explored for decades and keeps being researched; for a broader perspective, we refer the reader to the surveys on 3D reconstruction~\cite{berger2017survey}, RGB-D mapping~\cite{zollhofer2018state}, and neural fields~\cite{xie2022neural}. Mapping large 3D environments acquired by range scanners or LiDARs has been approached from a variety of perspectives. To achieve smooth and continuous surfaces, implicit surface representations such as surfels~\cite{pfister2000surfels}, volumetric truncated signed distance functions (TSDF)~\cite{curless1996volumetric,newcombe2011kinectfusion}, and implicit meshes~\cite{ilic2005implicit} are being actively integrated into reconstruction approaches. 
Here, TSDF representation was most extensively studied as it offers advantages for interpolation, multi-view fusion, and robustness to noise and topological changes.

TSDF fusion, starting with a classical approach~\cite{curless1996volumetric}, performs integration of a set of posed range-images into a coherent volumetric TSDF; memory requirements of the original method  scale linearly with the size of the output map. 
To extend TSDF fusion to large-scale 3D mapping such as outdoor LiDAR scans, its recent variants incorporate advanced spatial data structures such as hash tables~\cite{niessner2013real}, octrees~\cite{hornung2013octomap,steinbrucker2014volumetric,riegler2017octnetfusion}, and VDB~\cite{vizzo2022vdbfusion}.
In many instances, these methods provide robust reconstruction results, yet they may struggle to complete geometry in under-sampled areas as they lack  global geometric priors. 
To address the issue, recent scene completion approaches~\cite{dai2020sg,vizzo2022make} seek to predict occluded geometry in a self-supervised loop, going from incomplete to more complete fused reconstructions~\cite{curless1996volumetric,vizzo2022vdbfusion}.
Multiple methods explore combining volumetric occupancy and semantics for semantic scene completion (SSC)~\cite{yan2021sparse,cheng2021s3cnet,li2023voxformer,roldao2020lmscnet,xia2023scpnet} but require semantic annotations during training. 
Unlike our approach, all these methods deliver limited (albeit in some instances comparatively high) spatial resolution. 
We compare our method to recent volumetric TSDF fusion~\cite{vizzo2022vdbfusion} and learned completion~\cite{vizzo2022make} approaches.

Poisson Surface Reconstruction (PSR)~\cite{kazhdan2006poisson,kazhdan2013screened} is a seminal approach to implicit surface reconstruction from dense point clouds based on a local smoothness prior.
PUMA~\cite{vizzo2021poisson} extends PSR to offline mapping with sparse LiDAR 3D scans; we compare against this method in our evaluation. 

Multiple recent learning-based methods can learn effective reconstruction priors from shape collections.
For modelling closed, watertight shapes, pioneering approaches fit collections of latent codes and neural networks to datasets of point clouds, predicting signed distance functions (SDFs)~\cite{park2019deepsdf} or occupancy functions (OFs)~\cite{mescheder2019occupancy} as output.
Target objects with different properties may require choosing a different implicit representation; \eg, for open surfaces such as garments, sign-agnostic~\cite{atzmon2020sal} or unsigned~\cite{mullen2010signing,chibane2020neural,zhou2023learning} representations prove more effective than SDFs.
Shapes with rich internal structures benefit from fitting generalised representations~\cite{ye2022gifs} descibing spatial relationships between any two points, rather than a point and a surface. 
Due to their limited capacity, these methods cannot fit complex scenes with satisfactory accuracy.

Scaling neural implicit functions to large scenes can be achieved by introducing a collection of spatially latent codes instead of a single latent code.
In this direction, dense volumetric feature grids are the simplest option~\cite{jiang2020local,chabra2020deep,peng2020convolutional}, yet again scale linearly with scene complexity.
To optimize memory use and speed up mapping, several recent approaches define and optimize a multi-resolution feature grid~\cite{takikawa2021neural,yu2022monosdf,zhu2022nice}.
Multi-resolution latent hierarchies have been developed for LiDAR 3D mapping~\cite{zhong2023shine,song2024n,wiesmann2023locndf,shi2023accurate}; among these, we compare to a representative LiDAR-based neural mapping method~\cite{zhong2023shine}.

Most recently, building on the success of neural radiance fields~\cite{mildenhall2021nerf} and neural implicit surfaces~\cite{wang2021neus}, several methods adopted volumetric rendering to serve for optimizing the latent grid~\cite{wang2022go,azinovic2022neural,deng2023nerf,isaacson2023loner}.
We compare our method a recent approach targeting LiDAR acquisitions~\cite{deng2023nerf}.

\section{Method}
\label{sec:method}

\begin{figure}[t]
\centerline{
\includegraphics[
width=0.8\columnwidth,
trim=0 0 0 0, 
clip=True]{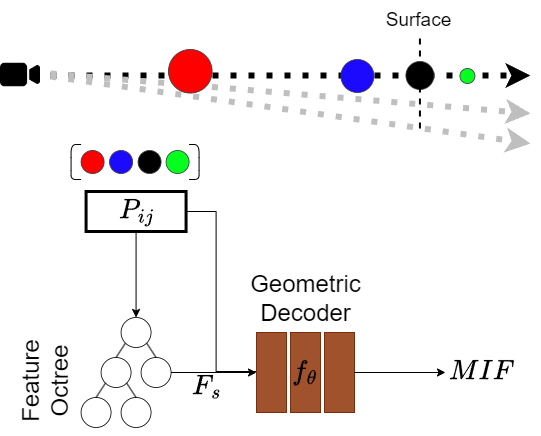}
}
\caption{Our algorithm comprises three main components: a sampling strategy, a feature octree, and an MLP decoder. 
For visualization purposes, each point is colored based on its input order and sized by its signed distance to the surface. Monotonicity loss is enforced according to this coloring order.}
\label{fig:method-overview}
\end{figure}

\subsection{Method Overview}
\label{method:overview}

Our algorithm accepts as input a set of posed 3D LiDAR scans $\{(P_i, T_i^L)\}_{i = 1}$ with 4$\times$4 LiDAR poses $T_i^L$. 
As output, it produces a 3D reconstruction of the scene in the form of a triangle mesh extracted from an (unknown) implicit field $f$ satisfying general conditions outlined in \Cref{method:overview}.
We approximate the field by a neural network $f_{\theta}$ (\Cref{method:architecture}) designing it similarly to existing neural implicit fields~\cite{park2019deepsdf,jiang2020local}; specifically, our implicit network accepts a feature vector~$\bm{z} \in \mathbb{R}^d$ and a 3D location~$\bm{x} \in \mathbb{R}^3$, and generates a value $y \in \mathbb{R}$ of a volumetric function via $y = f_{\theta}(\bm{x}; \bm{z})$.
To learn our neural field on the given data, we optimize a set of 3D losses imposed on our approximator $f_{\theta}$.
A high-level illustration of our pipeline is presented in \cref{fig:method-overview}.

\subsection{Monotonic Implicit Scene Representation}
\label{method:representation}

\paragraph{Neural Implicits and Level Sets}
Scenes can be represented by prescribing a scalar value (\eg, signed distance-to-surface~\cite{park2019deepsdf} or binary occupancy~\cite{mescheder2019occupancy}), $s \in \mathbb{R}$, to each 3D point sample~$\bm{x} \in \mathbb{R}^3$; then, an \textit{implicit function} $f(\bm{x}): \mathbb{R}^3 \to \mathbb{R}$ is said to describe the surface~$\mathcal{S}$ of the scene.
For instance, a (projective) signed distance function (SDF) maps 3D points to their closest projections on the (closed) surface via 
\begin{equation}
\label{eq:definition_of_sdf}
f_{\text{SDF}}(\bm{x}) = (-1)^{\chi_{\mathcal{V}}(\bm{x})} \cdot \arg \min\limits_{\bm{p} \in \mathcal{S}} ||\bm{x} - \bm{p}||
\end{equation}
where $\mathcal{V}$ is the volume enclosed by~$\mathcal{S}$ and $\chi_{\mathcal{V}}(\bm{x})$ its mem\-ber\-ship function.
To extract the surface of the scene explicitly (\eg, in the form of a triangle mesh), one computes a level-set $\{\bm{x}: f(\bm{x}) = s_0\}$ of~$f$ at a certain level~$s_0$ such as zero.

Implicit fields such as SDFs can be represented by neural nets~\cite{park2019deepsdf}; shape collections (or spatially large scenes~\cite{chabra2020deep}) can be encoded by parameterizing (conditioning) a learned implicit function~$f_{\theta}(\bm{x}; \bm{z})$ with a multivariate latent variable~$\bm{z} \in \mathbb{R}^d$ whose values are optimized jointly with $f_{\theta}$.

\paragraph{Our Implicit Representation}
Assuming that true values of the sought SDF in~\eqref{eq:definition_of_sdf} can be sampled at any given 3D point, conventional methods~\cite{park2019deepsdf,chabra2020deep} sample points exhaustively near surfaces to optimize approximator parameters~$\theta$ and a collection of latent shape codes~$\{z_i\}$.
For LiDAR 3D scans, SDF values are unknown for most 3D locations apart from a sparse set of points sampled along rays connecting scanned 3D points~$\{\bm{p}_i\}$ and scanner locations. 
Specifically, if~$\bm{o}$ is the sensor location and $\bm{q}(t) = \bm{o} + t \bm{d}_i$ is a laser ray emitted from the sensor in the direction~$\bm{d}_i$ towards point~$\bm{p}_i$, then the distance~$f_{\text{ray}}(\bm{q}) = ||\bm{q} - \bm{p}_i||$ encodes proximity to the surface. 
However, treating such values as \textit{projective distances} (in the sense of~\cref{eq:definition_of_sdf}) is incorrect as LiDAR rays are not orthogonal to surfaces; instead, $f_{\text{ray}}(\bm{q})$ encodes \textit{oblique distances} that significantly deviate from projective ones (\eg, at low incidence angles) and vary across sensor positions (see \cref{fig:method-oblique-sdf}).
Using oblique SDFs directly for learning yields imprecise, contradictory training signal if used with multiple aligned LiDAR scans, leading to blurry, incomplete reconstructions (\cref{fig:method-mif-vs-others}).

To circumvent the flaws of oblique SDFs, multiple methods transform them into occupancy probabilities by a sigmoid function (SHINE~\cite{zhong2023shine}), or project them along estimated surface normals (NeRF-LOAM~\cite{deng2023nerf} and N$^{\text{3}}$-Mapping~\cite{song2024n}) or along gradients of neural approximators (LocNDF~\cite{wiesmann2023locndf}). 
These methods are able to achieve success in some instances, but are still not free from limitations as neither of them compensates for non-projective distances completely. 
Estimating normals from sparse, noisy point clouds is known to be an unstable operation (see, \eg, \cite{koch2019abc}).

In this work, we adopt a different approach and let the network optimize a surface-aware implicit field without fitting all sampled distance-to-surface values exactly.
Instead, values of our \emph{generalized} implicit field are 
\begin{enumerate}
\item[(a)] zero at surface locations (\ie, lidar readings);
\item[(b)] positive outside and negative inside the shape;
\item[(c)] monotonically non-increasing along each cast ray. 
\end{enumerate}
We refer to our implicit function as \textit{monotonic implicit field (MIF)}. 
While conditions (a) and (b) are standard for implicit functions including SDFs and occupancy maps~\cite{park2019deepsdf,mescheder2019occupancy}, (c) aims to relax the requirement of precisely matching conflicting values of oblique distances corresponding to different scans. 
While exceptions to this assumption may occur (particularly, for rays passing near but outside shapes), we consider it to be more realistic compared to the assumption of correct SDF values. 
Hence, a wider range of non-metric implicit fields consistent with LiDAR data can be obtained (\cref{fig:method-mif-vs-others}). 
Our MIF can be meshed (\eg, by marching cubes~\cite{lewiner2003efficient}) as a usual implicit function to obtain a surface.

\subsection{Neural Architecture for 3D Mapping}
\label{method:architecture}

\begin{table}[b!]
\centerline{
\includegraphics[width=\columnwidth]{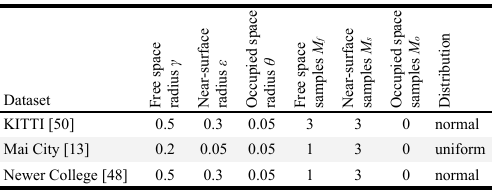}}
\caption{Best-performing sampling parameters for our method. 
Radii are given in meters. }
\label{tbl:implementation-sampling-details}
\end{table}


\begin{table*}[ht!]
\centerline{
\includegraphics[width=\textwidth]{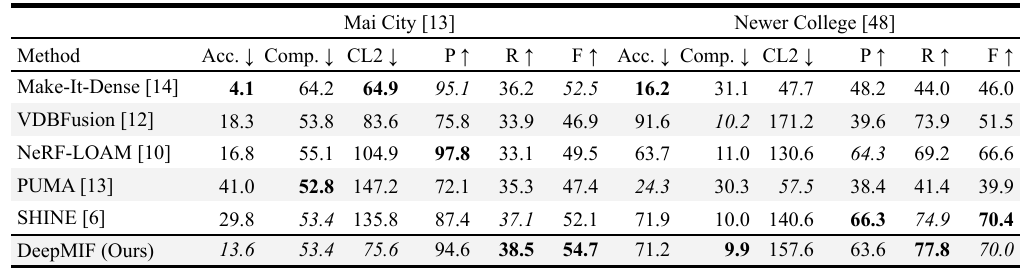}}
\caption{Quantitative comparison of reconstruction quality on Mai City~\cite{vizzo2021poisson} and Newer College~\cite{ramezani2020newer} benchmarks.
Values of Acc., Comp., CL2 are in centimeters;  values of P, R, F are in percentages; arrows indicate whether higher ($\uparrow$) or lower ($\downarrow$) values correspond to better results.  
The best performing method is presented in bold, the second best in italics.}
\label{tbl:comparative-quantitative-maicity-newercollege}
\end{table*}

\begin{table}[ht!]
\centerline{
\includegraphics[width=\columnwidth]{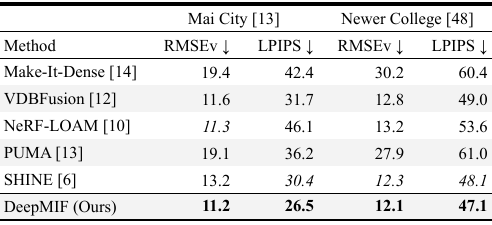}}
\caption{Quantitative evaluation of perceptual reconstruction quality across the Newer College and Mai City datasets. RMSE$_{\text{v}}$ and LPIPS metrics are provided for each method, with LPIPS calculated using a pretrained VGG backbone.
}
\label{tbl:comparative-perceptual-maicity-newercollege}

\end{table}


\paragraph{Query Point Sampling}
Let~$P_i = \{(\bm{p}_{ij}, \tau_{ij})\}$ be a point cloud acquired by a sensor emitting rays in directions~$\{\bm{d}_{ij}\}$ from~$\bm{o}_i$, where $\tau_{ij} = ||\bm{p}_{ij} - \bm{o}_i||$ refers to the acquired depth.
To produce training point instances, we sample points according to the relation~$\bm{p}^{\text{tr}}_{ij}(t) = \bm{o}_i + t \bm{d}_{ij}$ by selecting values of the parameter~$t$.
Specifically, we sample $M_{f}$, $M_{s}$ and $M_{o}$ values of~$t$ from segments $[\tau_{ij} - \gamma - \varepsilon, \tau_{ij} - \varepsilon]$, $[\tau_{ij} - \varepsilon, \tau_{ij} + \varepsilon]$, and $[\tau _{ij}+ \varepsilon, \tau_{ij} + \varepsilon + \theta]$, corresponding to outside-, near-, and within-surface samples, respectively. 
For training, we record pairs $(\bm{p}_{ij,m}, r_{ij,m})$ of point coordinates and signed distances $r_{ij,m} = \tau_{ij,m} - t_{ij}$ to the original readings, and collect the original sets~$P_i$ and sampled data into our final training set~$P$. 
Within each ray, generated points are sorted \wrt their values of $t$ and serve for direct supervision of our implicit function.
Samples within $\varepsilon$ to sensor readings are used for building the feature octree.

\paragraph{Hierarchical Feature Octree}
Our approach, similarly to existing frameworks~\cite{park2019deepsdf,jiang2020local}, jointly optimizes network parameters and local latent codes associated with sampled points during training.
More specifically, we take inspiration from SHINE~\cite{zhong2023shine} and assign latent codes to leaf nodes of a multi-resolution hierarchical octree constructed on top of the input point cloud. 
Constructing a local latent code with the octree hierarchy involves querying eight nodes from last $H$ level of the octree hierarchy, trilinearly interpolating them to form level-specific codes, and fusing level-specific codes into the aggregated latent code through summation. 
For constant-time queries, we convert points into locality-preserving spatial hashes (Morton codes) and query hash tables constructed for each level in the octree~\cite{zhong2023shine}. 
The point coordinates and its latent code are fed into the decoder network to predict the value of the implicit function.

\paragraph{Network Architecture}
Our network architecture follows the auto-decoder framework introduced in~\cite{park2019deepsdf}. 
We feed the geometric decoder, an MLP, with points sampled along the LiDAR ray by concatenating their corresponding feature vectors. Inspired by \cite{mildenhall2021nerf}, we incorporate positional encoding on the input points before concatanetion. 
To compute our monotonicity loss, we sort samples from the same ray according to the distance to the scanner before feeding into our network. 

\begin{figure*}[t]

\centering
\includegraphics[width=\textwidth]{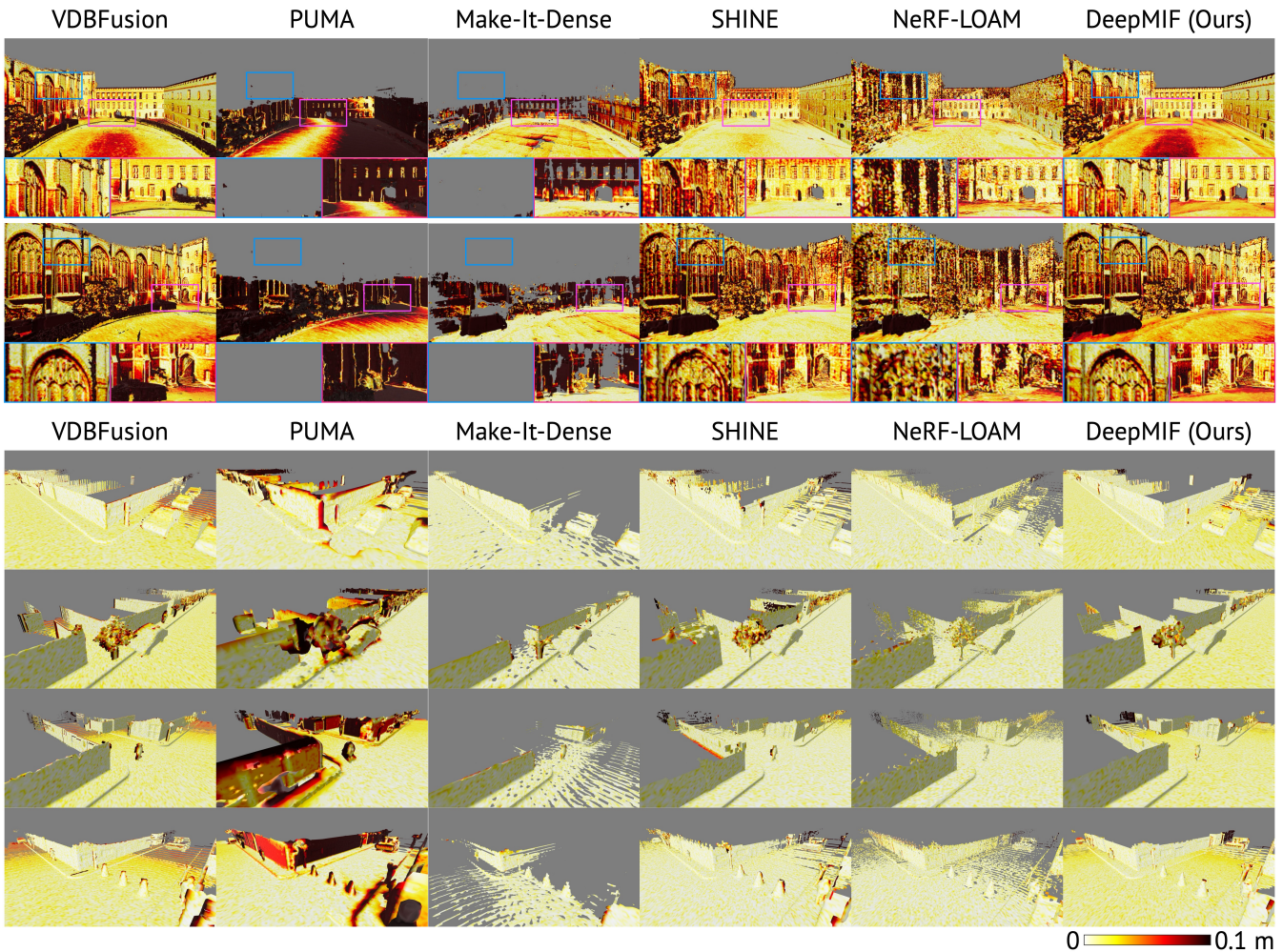}
\caption{\textit{Qualitative large-scale 3D mapping results on Newer College~\cite{ramezani2020newer} (upper part) and Mai City~\cite{vizzo2021poisson} (lower part).}
For Newer College, our algorithm delivers significantly cleaner reconstruction compared to NeRF-LOAM~\cite{deng2023nerf}, more complete results compared to PUMA~\cite{vizzo2021poisson} and Make-It-Dense~\cite{vizzo2022make},
and performs qualitatively comparably to VDBFusion~\cite{vizzo2022vdbfusion} and SHINE~\cite{zhong2023shine}. 
Similarly for Mai City, our algorithm obtains more complete, robust reconstructions, particularly at object edges.
\vspace{-1em}
}

\label{fig:experiments-comparative-gallery}
\end{figure*}

\paragraph{Losses}
We construct a neural approximator~$f_{\theta}$ of our implicit function using gradient descent to minimize a set of objective functions corresponding to its properties (\cref{method:representation}). 
To force our learned function to take zero values in surface 3D points (raw sensor readings), we minimize
\begin{equation}
\label{eq:loss_surf}
L_{\text{surf}} = \frac{1}{\lvert P_{\text{surf}} \rvert} \sum_{\bm{p} \in P_{\text{surf}}} | f_{\theta}(\bm{p}) |
\end{equation}
where $P_{\text{surf}} = \{\bm{p} | (\bm{p}, r) \in P, r = 0\}$ is the set of all such points. 
To explicitly encourage our function to produce values with a correct sign (\enquote{inside} or \enquote{outside} the surface) in each sampled point, we minimize
\begin{equation}
\label{eq:loss_sign}
L_{\text{sign}} = 
\frac{1}{|P|} \sum_{(\bm{p}, r) \in P} 
    \big(1 - s_{\bm{p}} \cdot l_{r} \big)
\end{equation}
where $s_{\bm{p}} = \sigma(f_{\theta}(\bm{p}))$ correspond to the \textit{soft sign} scores transforming the implicit function to a binary variable. 
Here, $\sigma(\alpha x)$ is a sigmoid function (we implement $\sigma$ using $\tanh$), where $\alpha > 0$ is a parameter controlling flatness of the function.
$l_r = \sigma(\alpha r)$ corresponds to the sigmoid-transformed signed distance~$r$ to sensor point computed along the ray.
Importantly, we aim to make our implicit function monotonically decreasing along LiDAR rays.
Specifically, for any two consecutive points $(\bm{p}_{ij,m}, r_{ij,m})$ and $(\bm{p}_{ij,m+1}, r_{ij,m+1})$ sampled on the same LiDAR ray $(i,j)$, the difference $\Delta_{ij,m} = f_{\theta}(\bm{p}_{ij,m}) - f_{\theta}(\bm{p}_{ij,m+1})$ should be positive.
Hence, we add a monotonicity objective  
\begin{equation}
\label{eq:loss_mono}
L_{\text{mono}} = 
\frac{1}{|\mathcal{R}(P)|} 
\sum_{(i,j) \in \mathcal{R}(P)}
\frac{1}{M} 
\sum_{m = 1}^{M} 
\big(1 - \delta_{ij,m}\big)
\end{equation}
where $\mathcal{R}(P) = \{(i,j)\}$ is the set of LiDAR rays emitted from all scanning positions and $\delta_{ij,m} = \sigma(\alpha \Delta_{ij,m})$ is the output of the sigmoid.
Finally, following prevailing practice~\cite{gropp2020implicit,yu2022monosdf,zhong2023shine}, we minimize the eikonal loss, enforcing the gradients on the implicit surface to be equal to 1:
\begin{equation}
L_{\text{eik}} =  \frac{1}{\lvert P_{\text{surf}} \rvert} 
\sum_{\bm{p} \in P_{\text{surf}}} (\| \nabla_{\bm{p}} f(\bm{p}) \|_2 - 1)^2.
\end{equation}
Our final geometric loss is given by
\begin{equation}
L_{\text{geo}} = L_{\text{surf}} + \lambda_{\text{eik}} L_{\text{eik}} + \lambda_{\text{sign}} L_{\text{sign}} + \lambda_{\text{mono}} L_{\text{mono}}
\end{equation}

\section{Experiments}
\label{sec:experiments}

\subsection{Experimental Setup}
\label{experiments:setup}

\begin{figure*}[hbt!]
\centerline{
\includegraphics[
width=\textwidth]{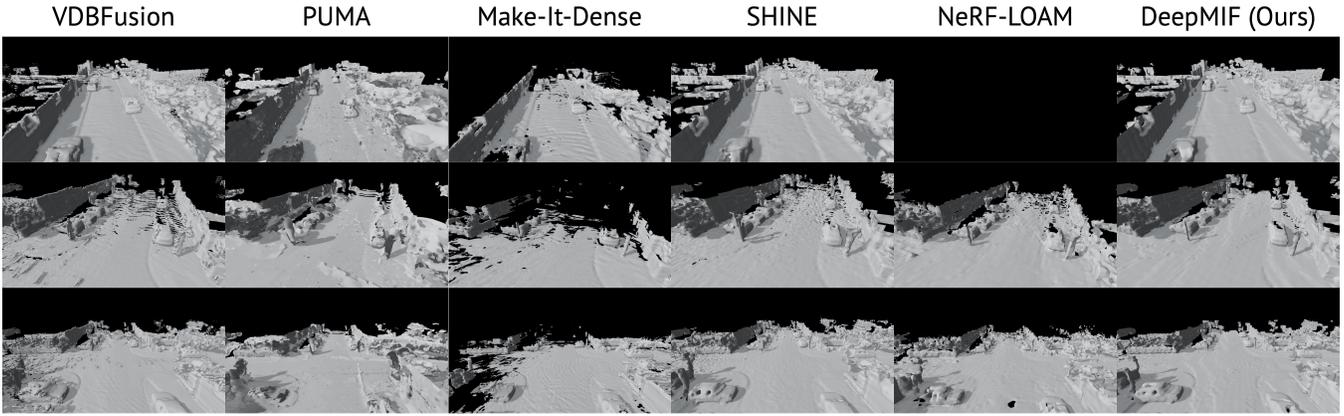}
}
\caption{\textit{Qualitative large-scale 3D mapping results on KITTI~\cite{geiger2012we}.} 
Compared to baselines, our method produces more complete, smooth, and sharp reconstruction.
\vspace{-1em}
}
\label{fig:experiments-kitti}
\end{figure*}

\begin{figure*}[hbt!]
\centerline{
\includegraphics[
width=\textwidth]{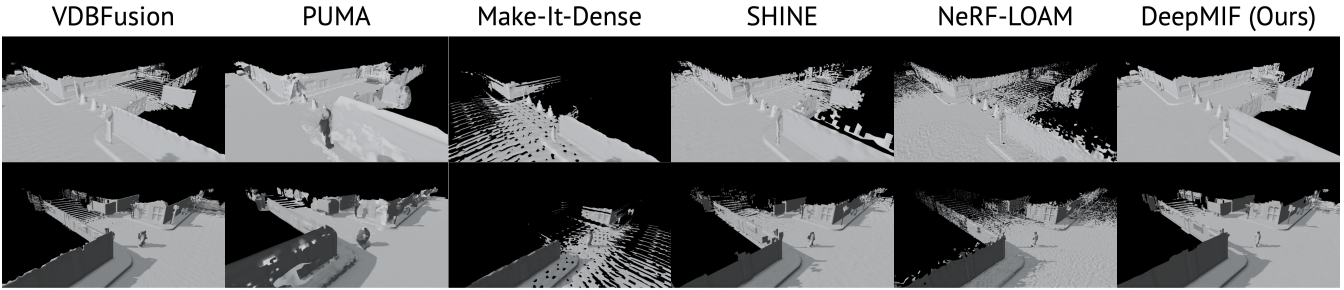}
}
\caption{\textit{Qualitative large-scale 3D mapping results on MaiCity~\cite{vizzo2021poisson}.} 
Compared to baselines, our method produces more complete, smooth, and sharp reconstruction.
\vspace{-1em}
}
\label{fig:experiments-maicity}
\end{figure*}

\paragraph{Baselines} 
We compare our method against a variety of state-of-the-art methods designed for large-scale outdoor LiDAR 3D mapping. 
SHINE~\cite{zhong2023shine} is a neural surface fitting method integrating a hierarchical latent grid. 
NeRF-LOAM~\cite{deng2023nerf} is a NeRF-based LiDAR 3D mapping method.
Make-It-Dense~\cite{vizzo2022make} is a learnable, self-supervised 3D scan completion method. 
PUMA~\cite{vizzo2021poisson} is a surface reconstruction method based on PSR~\cite{kazhdan2013screened}.
VDBFusion~\cite{vizzo2022vdbfusion} is a depth fusion method adapted to sparse LiDAR 3D scans.

\paragraph{Benchmark Datasets} 
We use three open-source datasets to evaluate our method. 
For quantitatively assessing 3D mapping performance, we use the simulated Mai City~\cite{vizzo2021poisson} benchmark, providing a single large-scale CAD scene and synthetic measurements obtained by a virtual 64-beam LiDAR scanning.
To quantify 3D mapping performance using real-world data, we use Newer College~\cite{ramezani2020newer}, a real-world dataset captured by a hand-held 64-beam LiDAR sensor and including a high-quality reference point cloud obtained by an industrial 3D laser scanner.
We additionally include qualitative results obtained on KITTI autonomous driving dataset including a 64-beam LiDAR scanner~\cite{geiger2012we}. 

To preprocess raw data, we filter LiDAR scans to only keep points within the range of 1.5\,m to 50\,m from the sensor. 
To support efficient processing, we downsample each LiDAR scan to a voxel resolution of 5\,cm. 
To identify and eliminate outliers in raw scans, we compute an average distance from each point to 25 its closest neighbors, removing points with a standard deviation in this quantity exceeding 2.5\,m.

\paragraph{Evaluation Metrics} 

Following standard practices~\cite{vizzo2021poisson,zhong2023shine}, we evaluate 3D surface reconstruction using common performance measures. Both the predicted and ground-truth meshes are uniformly sampled at a resolution of 2\,cm. Accuracy (Acc.) is the average distance between sampled points and the ground-truth mesh, while Completion (Comp.) measures the average distance from ground-truth points to the predicted mesh, truncated at 2\,m. Chamfer distance (CL2) is the squared mean of accuracy and completion. Precision (P) and Recall (R) are the ratios of points within a 10\,cm threshold from the predicted to the ground-truth mesh and vice versa, respectively. The F-Score (F) is the harmonic mean of Precision and Recall. We also assess the perceptual quality of 3D maps by comparing shaded renders to reference meshes, calculating perceptual~$\text{RMSE}_{\text{v}}~$\cite{voynov2019perceptual} and LPIPS~\cite{zhang2018unreasonable} measures, and reporting their averages.
For Mai City, we use 160 views generated from 20 poses along the road, each with 8 views around a $360^{\circ}$ circumference; for Newer College, 320 views were generated from 64 poses along the ground truth trajectory, with each pose elevated by 2 meters over 5 levels. These rendered views were compared with the ground truth mesh obtained from a surface reconstructed using the ground-truth point cloud.

\paragraph{Training Details} 
For positional encoding, we employ 10 periodic functions per point dimension (x, y, and z).
Our decoder is a 4-layer weight-normalized \cite{salimans2016weight} MLP with a hidden layer size of 256.
The feature vectors for each point are queried from the last 3 levels of a hierarchical octree, with each feature vector having a length of 8.
\cref{tbl:implementation-sampling-details} presents details on the sampling strategy used for each evaluation dataset.
We set the flatness coefficient $\alpha$ of the sigmoid function $\sigma(\alpha x)$ to 100.

We use the AdamW optimizer with a learning rate of 0.01, $\epsilon = 10^{-15}$, weight decay of $10^{-7}$, and perform learning rate decay steps at 10K and 50K iterations. 
Training employs batches of ray samples, where each input comprises multiple consecutive points along a LiDAR scan ray. 
This results in a total of $N \times B$ points processed per iteration, where $N$ is the number of samples and $B$ is the batch size. 
We train for 10K--20K iterations, taking approximately 30--60\,min on an NVIDIA RTX A5000 GPU and Intel Core i7 CPU.

\subsection{Comparisons to State-of-the-Art}
\label{experiments:comparative}

\paragraph{Quantitative Results}
\cref{tbl:comparative-quantitative-maicity-newercollege} shows a quantitative comparison of our method against baseline approaches at a reconstruction resolution of 10\,cm. 
While Make-It-Dense~\cite{vizzo2022make} achieves the highest accuracy, its reconstructions are incomplete and distorted (\cf~\cref{fig:experiments-comparative-gallery}), resulting in poor completion performance. 
Our method performs comparably to VDBFusion~\cite{vizzo2022vdbfusion} and SHINE~\cite{zhong2023shine} in both accuracy and completion, often achieving second place in Completion and F-Score. 
However, due to the limitations of the accuracy metric, which focuses only on predicted meshes, and the potential saturation of the completion metric by bloated predictions, point cloud comparisons may not fully reflect mesh smoothness or completion. 

Thus, we include a complementary evaluation in \cref{tbl:comparative-perceptual-maicity-newercollege} which provides an assessment of the perceptual performance of reconstruction methods on the Newer College and Mai City datasets using $\text{RMSE}_{\text{v}}$ and LPIPS metrics. These results show that our method outperforms others on both datasets across both metrics, indicating a closer visual resemblance to the ground truth meshes. Rendered view examples in \cref{fig:experiments-maicity} further demonstrate the superiority of our method in completion and reconstruction compared to baseline methods.

\paragraph{Qualitative Results on KITTI}
\cref{fig:experiments-kitti} presents a qualitative comparison of reconstruction methods on the KITTI dataset. 
Here, VDBFusion~\cite{vizzo2022vdbfusion} has strong object reconstruction performance but yields many missing areas and artifacts; SHINE~\cite{zhong2023shine} produces good reconstruction quality but features noisy, uneven, and incomplete surfaces. NeRF-LOAM~\cite{deng2023nerf} performs similarly to SHINE, yielding somewhat more incomplete reconstructions. 
Make-It-Dense~\cite{vizzo2022make} delivers high-quality reconstructions but lacks overall scene completion; in contrast, PUMA~\cite{vizzo2021poisson} focuses on scene completion but compromises on reconstruction quality and suffers from noticeable artifacts. 
Our method surpasses the others by effectively filling in missing parts and producing consistently smoother surfaces.

\subsection{Ablative Studies}
\label{experiments:ablative}

\cref{tbl:ablative-maicity-newercollege} provides results from an ablation study on the Mai City dataset, highlighting the impact of each loss term on reconstruction quality. 
Omitting the surface loss entirely results in reconstruction failure, suggesting training instability even though the combined sign and monotonicity losses implicitly contain surface information. 
Excluding either the monotonicity or sign loss produces similar results, but sign loss notably improves completion, while monotonicity loss enhances accuracy. 
Including all loss terms together yields the best overall performance across most metrics, except for completion, which remains comparable to the best-performing configuration. 
The eikonal loss was excluded from this study to maintain focus on the specific effects of the other loss terms.




\begin{table}[t!]
\centerline{
\includegraphics[width=\columnwidth]{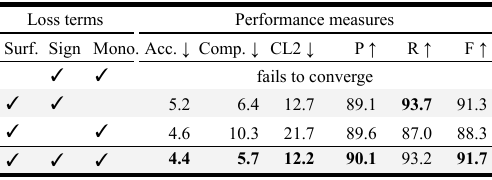}}
\caption{Contribution of individual loss terms into overall performance. 
Values of Acc., Comp., CL2 are in centimeters;  values of P, R, F are in percentages; arrows indicate whether higher ($\uparrow$) or lower ($\downarrow$) values correspond to better results. }
\label{tbl:ablative-maicity-newercollege}
\end{table}


\section{Conclusion}
\label{sec:conclusion}

We proposed a new implicit representation suitable for LiDAR-based 3D scene modelling.
Compared to existing representations such as signed distance functions, our monotonic implicit function does not require exact, dense point samples to be trained from sparse point sets acquired by modern scanners such as LiDARs.
Our implicit field can be easily integrated in a large-scale 3D mapping system by enforcing a monotonicity loss along sensor's scanning rays. 
We have demonstrated the capabilities of our method with a synthetic Mai City and a real-world Newer College and KITTI benchmarks, where we achieved strong performance compared to five distinct baseline approaches.

{\small
\bibliographystyle{IEEEtran}
\bibliography{IEEEabrv,references}
 }

\end{document}